\pdfoutput=1

\documentclass{article}

\usepackage{microtype}
\usepackage{graphicx}
\usepackage{booktabs} 

\usepackage{hyperref}
\usepackage{cleveref}

\usepackage[accepted]{icml2019}

\usepackage[]{amsmath}
\usepackage{macros}

\usepackage{subcaption}
\usepackage{xcolor}
\usepackage{bm}
\usepackage{tikz}
\usetikzlibrary{shapes, arrows, positioning, fit, calc, arrows.meta}
\usetikzlibrary{bayesnet, positioning}
\tikzset{>=latex}
\usepackage{mathtools}
\usepackage{natbib}
\usepackage{mwe}
\usepackage{adjustbox}
\usepackage{array}

\tikzset{%
  var/.style      = {draw, thick, circle, minimum size=0.6cm, inner sep=0},
  block/.style    = {draw, thick, rectangle, rounded corners},
  customline/.style    = {draw opacity=0.4, line width=1}
}

\definecolor{phicolor}{HTML}{0074D9}
\definecolor{gammacolor}{HTML}{FF851B}

\icmltitlerunning{Autoregressive Energy Machines}

\begin{document}

\twocolumn[
\icmltitle{Autoregressive Energy Machines}



\icmlsetsymbol{equal}{*}

\begin{icmlauthorlist}
\icmlauthor{Charlie Nash}{equal,ed}
\icmlauthor{Conor Durkan}{equal,ed}

\end{icmlauthorlist}

\icmlaffiliation{ed}{School of Informatics, University of Edinburgh, United Kingdom}

\icmlcorrespondingauthor{Charlie Nash}{charlie.nash@ed.ac.uk}
\icmlcorrespondingauthor{Conor Durkan}{conor.durkan@ed.ac.uk}

\icmlkeywords{Machine Learning, ICML}

\vskip 0.3in
]



\printAffiliationsAndNotice{\icmlEqualContribution} 

\begin{abstract}
Neural density estimators are flexible families of parametric models which have seen widespread use in unsupervised machine learning in recent years. Maximum-likelihood training typically dictates that these models be constrained to specify an explicit density. However, this limitation can be overcome by instead using a neural network to specify an energy function, or unnormalized density, which can subsequently be normalized to obtain a valid distribution. The challenge with this approach lies in accurately estimating the normalizing constant of the high-dimensional energy function. We propose the Autoregressive Energy Machine, an energy-based model which simultaneously learns an unnormalized density and computes an importance-sampling estimate of the normalizing constant for each conditional in an autoregressive decomposition. The Autoregressive Energy Machine achieves state-of-the-art performance on a suite of density-estimation tasks.
\end{abstract}

\section{Introduction}\label{sec:introduction}
Modeling the joint distribution of high-dimensional random variables is a key task in unsupervised machine learning. In contrast to other unsupervised approaches such as variational autoencoders \cite{kingma2013vae, rezende2014stochastic} or generative adversarial networks \cite{goodfellow2014gan}, neural density estimators allow for exact density evaluation, and have enjoyed success in modeling natural images \cite{oord2016pixelcnn, dinh2016realnvp, salimans2017pixelcnn++, kingma2018glow}, audio data \cite{van2016wavenet, prenger2018waveglow, kim2018flowavenet}, and also in variational inference \cite{rezende2015normalizing, kingma2016iaf}. Neural density estimators are particularly useful where the focus is on accurate density estimation rather than sampling, and these models have seen use as surrogate likelihoods \cite{papamakarios2018sequential} and approximate posterior distributions \cite{papamakarios2016fast, lueckmann2017flexible} for likelihood-free inference.

  \begin{figure}[H]
    \centering
    \begin{minipage}[b]{0.3\columnwidth}
      \begin{subfigure}[b]{\linewidth}
            \centering
            \includegraphics[width=\columnwidth]{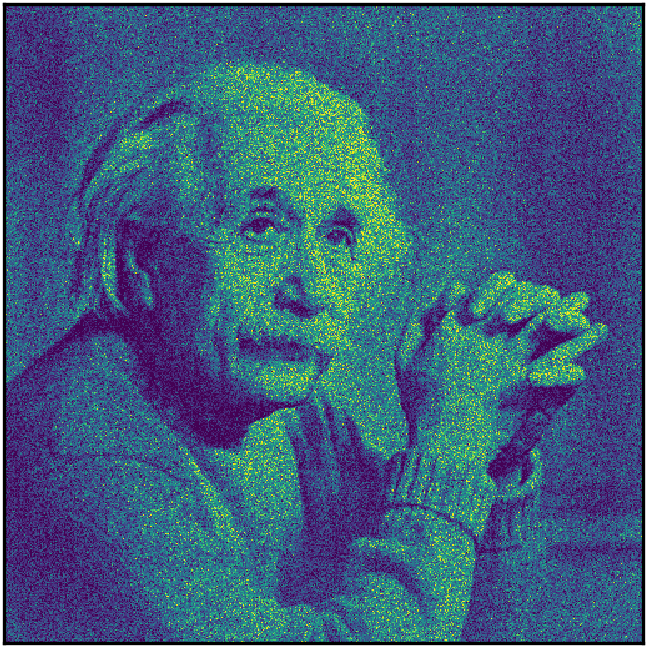}
        \subcaption{Data}
      \end{subfigure}
      \begin{subfigure}[b]{\linewidth}
            \centering
            \includegraphics[width=\columnwidth]{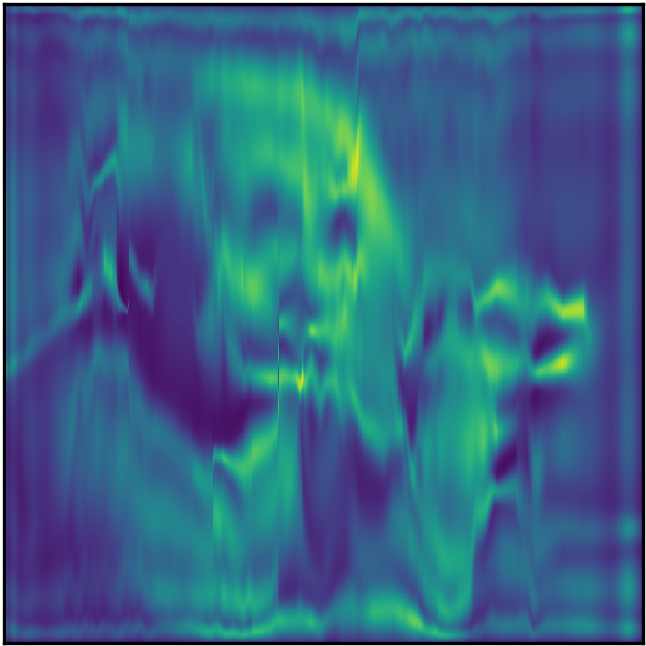}
        \subcaption{ResMADE}
      \end{subfigure}
    \end{minipage}
    \hfill
    \begin{subfigure}[b]{0.68\columnwidth}
        \centering
        \includegraphics[width=\columnwidth]{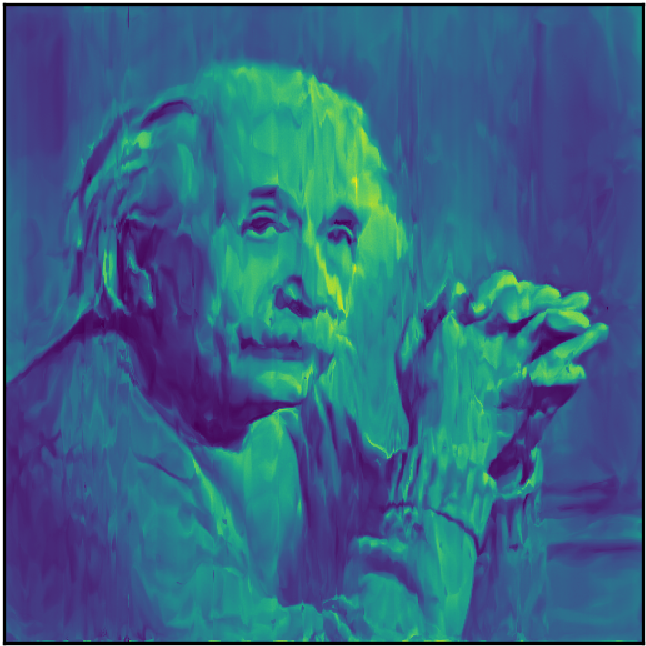}
        \vspace{-1.26em}
      \subcaption{AEM}
    \end{subfigure}
    \caption{Accurately modeling a distribution with sharp transitions and high-frequency components, such as the distribution of light in a image (a), is a challenging task. We find that an autoregressive energy-based model (c) is able to preserve fine detail lost by an alternative model (b) with explicit conditionals.}
    \label{fig:teaser}
  \end{figure}

Neural networks are flexible function approximators, and promising candidates to learn a probability density function. Typically, neural density models are normalized a priori, but this can hinder flexibility and expressiveness. For instance, many flow-based density estimators \cite{dinh2016realnvp, papamakarios2017maf, huang2018naf} rely on invertible transformations with tractable Jacobian which map data to a simple base density, so that the log probability of an input point can be evaluated using a change of variables. Autoregressive density estimators \cite{uria2013rnade, germain2015made} often rely on mixtures of parametric distributions to model each conditional. Such families can make it difficult to model the low-density regions or sharp transitions characterized by multi-modal or discontinuous densities, respectively.

The contributions of this work are shaped by two main observations.
\begin{itemize}
    \item An energy function, or unnormalized density, fully characterizes a probability distribution, and neural networks may be better suited to learning such an energy function rather than an explicit density.
    \item Decomposing the density estimation task in an autoregressive manner makes it possible to train such an energy-based model by maximum likelihood, since it is easier to obtain reliable estimates of normalizing constants in low dimensions. 
\end{itemize}

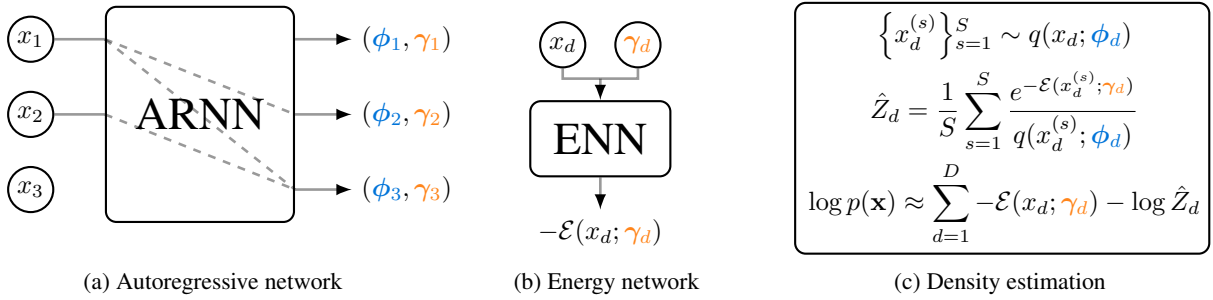
\begin{figure*}[t]
    \centering
    \begin{subfigure}[b]{0.32\textwidth}
        \centering
{
    \begin{tikzpicture}[auto, thick, node distance=1cm]
    
        \draw
        node [var] (x_1) {$x_1$}
    	node [var, below of=x_1] (x_2) {$x_2$}
    	node [var, below of=x_2] (x_3) {$x_3$}
    	
    	node [block, fit=(x_1)(x_3), xshift=2.25cm, minimum width=2.5cm, label=center:{\LARGE ARNN}] (AR-NN) {};
    
        \draw (5, 0) node (AR_outputs_1) {$(\textcolor{phicolor}{\bfphi_1}, \textcolor{gammacolor}{\bfgamma_1})$}
        node [below of=AR_outputs_1] (AR_outputs_2) {$(\textcolor{phicolor}{\bfphi_2}, \textcolor{gammacolor}{\bfgamma_2})$}
        node [below of=AR_outputs_2] (AR_outputs_3) {$(\textcolor{phicolor}{\bfphi_3}, \textcolor{gammacolor}{\bfgamma_3})$};
    
    	\draw[customline, -] (x_1) -- (x_1-|AR-NN.west);
    	\draw[customline, -] (x_2) -- (x_2-|AR-NN.west);
    
    	\draw[customline, dashed] (x_1-|AR-NN.west) -- (AR_outputs_2-|AR-NN.east);
    	\draw[customline, dashed] (x_1-|AR-NN.west) -- (AR_outputs_3-|AR-NN.east);
    	\draw[customline, dashed] (x_2-|AR-NN.west) -- (AR_outputs_3-|AR-NN.east);
    	
    	\draw[customline, ->] (AR_outputs_1-|AR-NN.east) -- (AR_outputs_1);
    	\draw[customline, ->] (AR_outputs_2-|AR-NN.east) -- (AR_outputs_2);
    	\draw[customline, ->] (AR_outputs_3-|AR-NN.east) -- (AR_outputs_3);
    	
    \end{tikzpicture}
}
        \caption{Autoregressive network}
        \label{fig:arnn}
    \end{subfigure}
    \begin{subfigure}[b]{0.28\textwidth}
        \centering
{
    \begin{tikzpicture}[auto, thick, node distance=1cm]
    
        \draw node [var] (x_d) {$x_d$};
    	\draw node [var, right of=x_d] (gamma_d) {$\textcolor{gammacolor}{\bfgamma_d}$};
    	
    	\draw node [block, fit=(x_d)(gamma_d), yshift=-1.25cm, minimum height=1cm, label=center:{\LARGE ENN}] (ENN) {};
    	 
    	\draw node [below of=ENN, yshift=-0.25cm] (energy) {$-\fancyE (x_d ; \textcolor{gammacolor}{\bfgamma_d})$};
    	
    	\draw node [below of=x_d, yshift=0.55cm, xshift=0.5cm] (hidden_node) {};
    	
    	\draw node [below of=hidden_node, yshift=-1cm] (space_node) {};

    	\draw[customline, -] (x_d) |- (hidden_node.center);
    	\draw[customline, -] (gamma_d) |- (hidden_node.center);
    	\draw[customline, ->] (hidden_node.center) -- (ENN.north);
    	
    	\draw[customline, ->] (ENN.south) -- (energy);

    \end{tikzpicture}
}
        \caption{Energy network}
        \label{fig:enn}
    \end{subfigure}
    \begin{subfigure}[b]{0.32\textwidth}
        \centering
        \begin{tikzpicture}[auto, thick, node distance=1cm]
\draw 
    node [block] () {%
    $ \begin{gathered}
    \curlybr{ x_{d}^{(s)} }_{s=1}^{S} \sim q (x_{d} ; \textcolor{phicolor}{\bfphi_{d}}) \\ 
    \Hat{Z}_{d} = \frac{1}{S} \sum_{s=1}^{S} \frac{e^ { - \fancyE(x_{d}^{(s)} ; \textcolor{gammacolor}{\bfgamma_{d}} ) }}{ q (x_{d}^{(s)}; \textcolor{phicolor}{\bfphi_{d}})} \\
    \log p(\bfx) \approx \sum_{d=1}^{D} -\fancyE({x_{d} ; \textcolor{gammacolor}{\bfgamma_{d}}}) - \log \Hat{Z}_{d} \\
\end{gathered} $};
\end{tikzpicture}
        \caption{Density estimation}
        \label{fig:equations}
    \end{subfigure}
    \caption{Overview of an AEM. (\protect\subref*{fig:arnn}) An autoregressive neural network computes an autoregressive function of an input $ \bfx $, such that the \dth output depends only on $ \bfx_{<d} $. The \dth output is a pair of vectors $ (\textcolor{phicolor}{\bfphi_{d}}, \textcolor{gammacolor}{\bfgamma_{d}}) $ which correspond to the proposal parameters and context vector, respectively, for the \dth conditional distribution. (\protect\subref*{fig:enn}) The context vector $ \textcolor{gammacolor}{\bfgamma_{d}} $ and input $ x_{d} $ are passed through the energy network to compute an unnormalized log probability for the \dth conditional. (\protect\subref*{fig:equations}) The parameters $ \textcolor{phicolor}{\bfphi_{d}} $ define a tractable proposal distribution, such as a mixture of location-scale family distributions, which can be used to compute an estimate of the normalizing constant $ \Hat{Z}_{d} $ for the \dth conditional by importance sampling.}
    \label{fig:aem} 
\end{figure*}

Based on these observations, we present a scalable and efficient learning algorithm for an autoregressive energy-based model, which we term the \textit{Autoregressive Energy Machine} (AEM\@). \Cref{fig:aem} provides a condensed overview of how an AEM approximates the density of an input point.   

\begin{figure}
	\centering
	\includegraphics[width=3in]{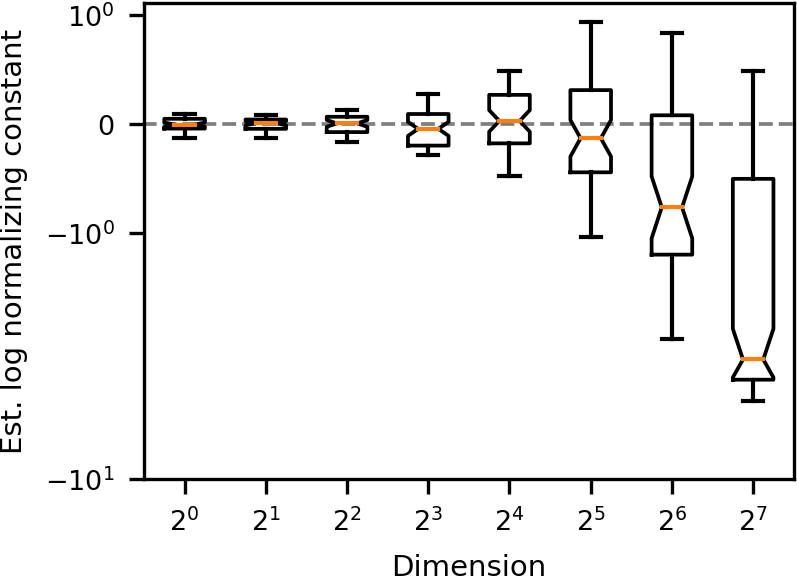}
	\caption{Importance sampling estimates of log normalizing constants deteriorate with increasing dimension. The target and proposal distributions are spherical Gaussians with $\sigma = 1$ and $\sigma = 1.25$, respectively. The true log normalizing constant is $\log{Z} = 0$. We plot the distribution of estimates over 50 trials, with each trial using 20 importance samples.}
	\label{fig:importance_sampling_scaling}
\end{figure}

\section{Background}\label{sec:background}

\subsection{Autoregressive neural density estimation}\label{subsec:neural_density_estimation}
A probability density function assigns a non-negative scalar value $ p(\bfx) $ to each vector-valued input $ \bfx $, with the property that $ \int p(\bfx) \diff \bfx = 1 $ over its support. Given a dataset $ \curlyD = \curlybr{\bfx^{(n)}}_{n=1}^{N} $ of $ N $ i.i.d.~samples drawn from some unknown $D$-dimensional distribution $ p^{\star}(\bfx) $, the density estimation task is to determine a model $  p(\bfx) $ such that $ p(\bfx) \approx p^{\star}(\bfx) $. Neural density estimators are parametric models that make use of neural network components to increase their capacity to fit complex distributions, and autoregressive neural models are among the best performing of these.

The product rule of probability allows us to decompose any joint distribution $ p(\bfx) $ into a product of conditional distributions:
\begin{align}
    p(\bfx) = \prod_{d=1}^{D} p(x_{d} \vert \bfx_{<d}).
    \label{eq:autoregressive}
\end{align} 
Autoregressive density estimators model each conditional using parameters which are computed as a function of the preceding variables in a given ordering. In this paper, we use the term ARNN to describe any autoregressive neural network which computes an autoregressive function of a $ D $-dimensional input $ \bfx $, where the \dth output is denoted $ \bff (\bfx_{<d}) $. The vector $ \bff (\bfx_{<d}) $ is often interpreted as the parameters of a tractable parametric density for the \dth conditional, such as a mixture of location-scale distributions or the probabilities of a categorical distribution, but it is not restricted to this typical use-case.



Certain architectures, such as those found in recurrent models, perform the autoregressive computation sequentially, but more recent architectures exploit masking or causal convolution in order to output each conditional in a single pass of the network. Both types of architecture have found domain-specific \cite{sundermeyer2012, theis2015generative, parmar2018imagetransformer}, as well as general-purpose \cite{uria2013rnade, germain2015made} use. In particular we highlight MADE \cite{germain2015made}, an architecture that masks weight matrices in fully connected layers to achieve causal structure. It is a building block in many models with autoregressive components \cite{kingma2016iaf, papamakarios2017maf, huang2018naf}, and we make use of a similar architecture in this work.

\subsection{Energy-based models}\label{subsec:energy_based_models}
In addition to an autoregressive decomposition, we may also write any density $ p(\bfx) $ as
\begin{align}
    p(\bfx) = \frac{e^{-\fancyE(\bfx)}}{Z},	
    \label{eq:energy}
\end{align}
where $ e^{-\fancyE(\bfx)} $ is the unnormalized density, $ \fancyE(\bfx) $ is known as the energy function, and $ Z = \int e^{-\fancyE(\bfx)} \diff \bfx $ is the normalizing constant. Assuming $ Z $ is finite, specifying an energy function is equivalent to specifying a probability distribution, since the normalizing constant is also defined in terms of $ \fancyE(\bfx) $. Models described in this way are known as energy-based models. Classic examples include Boltzmann machines \cite{hinton2002product, salakhutdinov09deep, hinton2012practical}, products of experts \cite{hinton2002product} and Markov random fields \cite{osindero2007patches, koster2009estimating}.

In order to do maximum-likelihood estimation of the parameters of an energy-based model, we must be able to evaluate or estimate the normalizing constant $ Z $. This is problematic, as it requires the estimation of a potentially high-dimensional integral. As such, a number of methods have been proposed to train energy-based models, which either use cheap approximations of the normalizing constant, or side-step the issue entirely. These include contrastive divergence \cite{hinton2002product}, noise-contrastive estimation \cite{gutmann2010noise, ceylan2018conditional}, and score matching \cite{hyvarinen2005estimation}. In our case, phrasing the density estimation problem in an autoregressive manner allows us to make productive use of importance sampling, a stochastic approximation method for integrals. 




\boldpar{Importance sampling} 
Given a proposal distribution $ q(\bfx) $ which is non-zero whenever the target $ p(\bfx) \propto e^{- \fancyE(\bfx)} $ is non-zero, we can approximate the normalizing constant $ Z $ by
\begin{align}
Z &= \int e^{-\fancyE(\bfx)} \diff \bfx = \int \frac{e^{-\fancyE(\bfx)}}{q(\bfx)} q(\bfx) \diff \bfx \\ 
&\approx \frac{1}{S} \sum_{s=1}^{S} \frac{e^{-\fancyE \roundbr{ \bfx^{(s)} }}}{q (\bfx^{(s)})}, \hspace{0.2cm} \bfx^{(s)} \sim q(\bfx), \label{eq:importance_weights}
\end{align}
and this expression is an unbiased estimate of the normalizing constant for $ p(\bfx) $. The quotients in the summand are known as the importance weights. When $ q(\bfx) $ does not closely match $ p(\bfx) $, the importance weights will have high variance, and the importance sampling estimate will be dominated by those terms with largest weight. Additionally, when $ q(\bfx) $ does not adequately cover regions of high density under $ p(\bfx) $, importance sampling underestimates the normalizing constant \cite{salakhutdinov2008quantitative}. Finding a suitable distribution $ q(\bfx) $ which closely matches $ p(\bfx) $ is problematic, since estimating the potentially complex distribution $ p(\bfx) $ is the original problem under consideration. This issue is exacerbated in higher dimensions, and importance sampling estimates may be unreliable in such cases. Figure \ref{fig:importance_sampling_scaling} demonstrates how the accuracy of an importance sampling estimate of the normalizing constant for a standard normal distribution deteriorates as dimensionality increases.

\section{Autoregressive Energy Machines}\label{sec:aem}

The ability to specify a probability distribution using an energy function is enticing, since now a neural network can take on the more general role of an energy function in a neural density estimator. Further decomposing the task in an autoregressive manner means that importance sampling offers a viable method for maximum likelihood training, generally yielding reliable normalizing constant estimates in the one-dimensional case when the proposal distribution is reasonably well matched to the target (\cref{fig:importance_sampling_scaling}). 
As such, the main contribution of this paper is to combine eq.~\eqref{eq:autoregressive} and eq.~\eqref{eq:energy} in the context of a neural density estimator. We model a density function $ p(\bfx) $ as a product of $ D $ energy terms
\begin{align}
    p(\bfx) &= \prod_{d=1}^{D} p(x_{d} \vert \bfx_{<d}) = \prod_{d=1}^{D} \frac{e^{-\fancyE(x_{d}; \bfx_{<d})}}{{Z_{d}}},
\end{align}
where $ Z_d = \int e^{-\fancyE(x_d ; \bfx_{<d})} \diff \bfx_d $ is the normalizing constant for the \dth conditional. If we also specify an autoregressive proposal distribution $q(\bfx) = \prod_{d} q(x_{d} \vert \bfx_{<d})$, we can estimate the normalizing constant for each of the $ D $ terms in the product by importance sampling: 
\begin{align}
    Z_{d} &= \int e^{-\fancyE(x_{d}; \bfx_{<d})} \diff x_{d} \\
    &\approx \frac{1}{S} \sum_{s=1}^{S} \frac{e^{-\fancyE\roundbr{x^{(s)}_{d}; \bfx_{<d}}}}{q(x^{(s)}_{d}; \bfx_{<d})}, \quad x^{(s)}_{d} \sim q \roundbr{x_{d}; \bfx_{<d}}.
\end{align}

This setup allows us to make use of arbitrarily complex energy functions, while relying on importance sampling to estimate normalizing constants in one dimension; a much more tractable problem than estimation of the full $ D $-dimensional integral.

\subsection{A neural energy function}
We implement the energy function as a neural network ENN that takes as input a scalar $x_d$ as well as a context vector $\bfgamma_d$ that summarizes the dependence of $x_d$ on the preceding variables $\bfx_{<d}$. The ENN directly outputs the negative energy, so that $ -\fancyE(x_{d}; \bfx_{<d}) = -\fancyE(x_{d}; \bfgamma_{d}) = \text{ENN}(x_{d}; \bfgamma_{d}) $. In our experiments, the ENN is a fully-connected network with residual connections \citep{he2016deep}. To incorporate the context vector $\bfgamma_d$, we found concatenation to the input $ x_{d} $ to work well in practice. We share ENN parameters across dimensions, which reduces the total number of parameters, and allows us to learn features of densities which are common across dimensions. We also constrain the output of the ENN to be non-positive using a softplus non-linearity, so that the unnormalized density is bounded by one, since this improved training stability.  



\subsection{Learning in an AEM}
We denote by $ \bfphi_{d} $ the vector of parameters for the \dth proposal conditional, which, like the context vector $ \bfgamma_{d} $, is computed as a function of $ \bfx_{<d} $. In our case, this quantity consists of the mixture coefficients, locations, and scales of a tractable parametric distribution, and we find that a mixture of Gaussians works well across a range of tasks. The normalizing constant for the \dth conditional can thus be approximated by
\begin{align}
    \hat{Z}_{d} = \frac{1}{S} \sum_{s=1}^{S} \frac{e^{-\fancyE\roundbr{x^{(s)}_{d}; \bfgamma_{d}}}}{q(x^{(s)}_{d}; \bfphi_{d})}, \quad x^{(s)}_{d} \sim q \roundbr{x_{d}; \bfphi_{d}},
\end{align}
leading to an expression for the approximate log density of an input data point $ \bfx $
\begin{align}
    \log p(\bfx) \approx \sum_{d=1}^D -\fancyE(x_d; \bfgamma_d) - \log \hat{Z}_d.
    \label{aem-log-prob}
\end{align}
Estimation of log densities therefore requires the energy network to be evaluated $S + 1$ times for each conditional; $S$ times for the importance samples, and once for the data point $x_d$. In practice, we perform these evaluations in parallel, by passing large batches consisting of input data and importance samples for all conditionals to the energy net along with the relevant context vectors, and found $ S = 20 $ to be sufficient. Although the importance sampling estimates of the normalizing constants are unbiased, by taking the logarithm of $\hat{Z}_d$ in eq.~\eqref{aem-log-prob} we bias our estimates of $\log p(\bfx)$. However, as we will show in Section \ref{subsec:well-calibrated}, our estimates are well-calibrated, and can be made more accurate by increasing the number of importance samples. For the purposes of training, we did not find this bias to be an issue.

As illustrated in \cref{fig:arnn}, we obtain both context vectors and proposal parameters in parallel using an autoregressive network ARNN\@. The ARNN outputs proposal parameters $ \bfphi_{d} $ and context vectors $ \bfgamma_{d} $ for each of the $ D $ dimensions in a single forward pass. These quantities are used both to estimate $ \log p(\bfx) $ as in eq.~\eqref{aem-log-prob}, as well as to evaluate $ \log q(\bfx) $, and we form a maximum-likelihood training objective
\begin{align}
    \mathcal{L}(\bftheta ; \bfx) =  \log p(\bfx) + \log q(\bfx) ,
    \label{eq:objective}
\end{align}
where $ \bftheta $ refers collectively to the trainable parameters in both the ARNN and the ENN\@. We fit the AEM by maximizing eq.~\eqref{eq:objective} across a training set using stochastic gradient ascent, and find that a warm-up period where the proposal is initially optimized without the energy model can improve stability, allowing the proposal to cover the data sufficiently before importance sampling begins. 

It is important to note that we do not optimize the proposal distribution parameters $ \bfphi $ with respect to the importance sampling estimate. This means that the proposal is trained independently of the energy-model, and estimates of $ \log p(\bfx) $ treat the proposal samples and proposal density evaluations as constant values for the purposes of optimization. In practice, this is implemented by stopping gradients on variables connected to the proposal distribution in the computational graph. We find maximum-likelihood training of $q$ to be effective as a means to obtain a useful proposal distribution, but other objectives, such as minimization of the variance of the importance sampling estimate \cite{kuleshov2017neural, muller2018neuralimportancesampling}, might also be considered, although we do not investigate this avenue in our work. 

\newcommand{\plotsizeplane}{0.75in}
\renewcommand\arraystretch{0.5}
\begin{figure*}[t]
    \begin{subfigure}[b]{0.601\textwidth}
    	\centering
    	\setlength\tabcolsep{1pt} 
    	\begin{tabular}{ccccc}
    	     & \multicolumn{2}{c}{Proposal} & \multicolumn{2}{c}{AEM} \\
    	    \cmidrule(lr){2-3} \cmidrule(lr){4-5}
    	    Data & Density & Samples & Density & Samples \\
        	\includegraphics[width=\plotsizeplane]{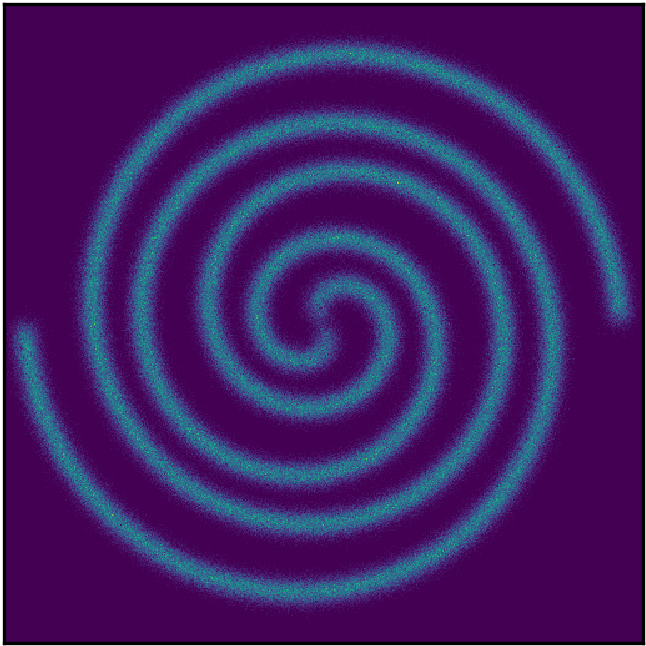} &
        	\includegraphics[width=\plotsizeplane]{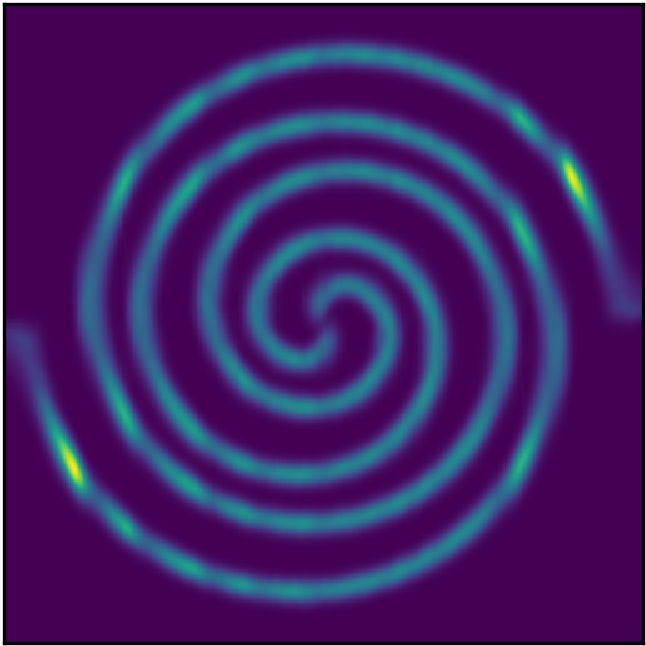} &
        	\includegraphics[width=\plotsizeplane]{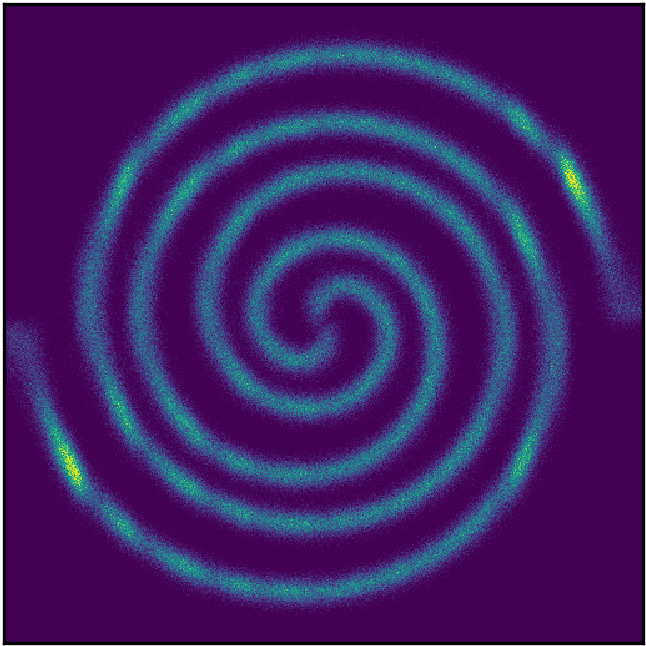} &
        	\includegraphics[width=\plotsizeplane]{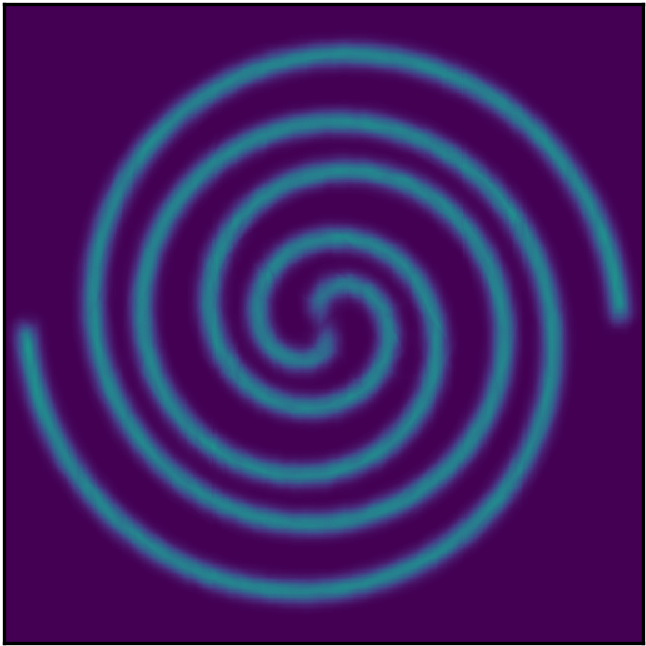} &
        	\includegraphics[width=\plotsizeplane]{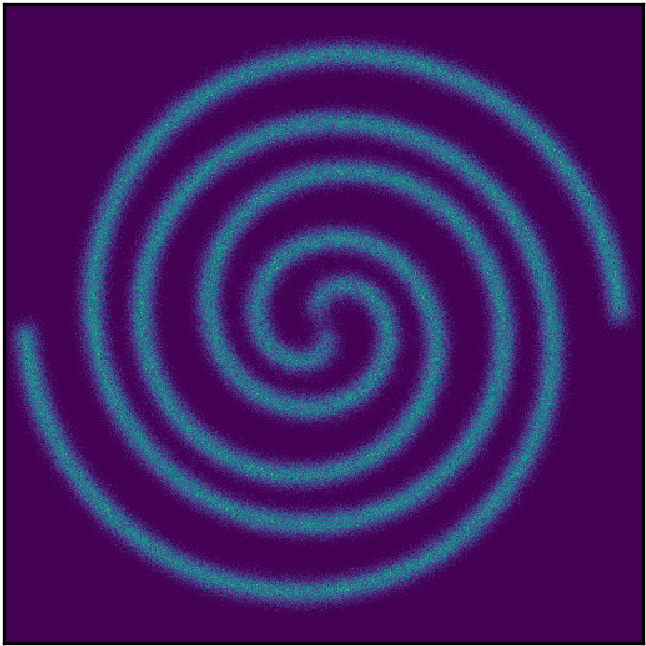}
        	\\
        	\includegraphics[width=\plotsizeplane]{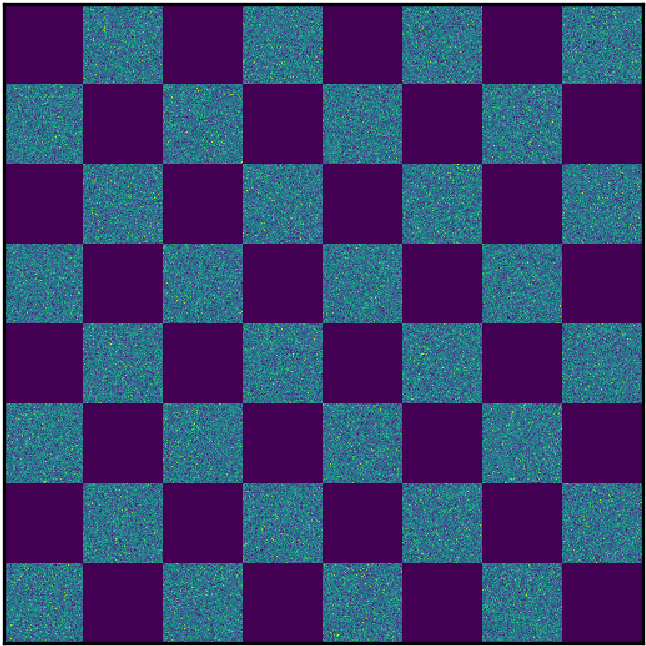} &
        	\includegraphics[width=\plotsizeplane]{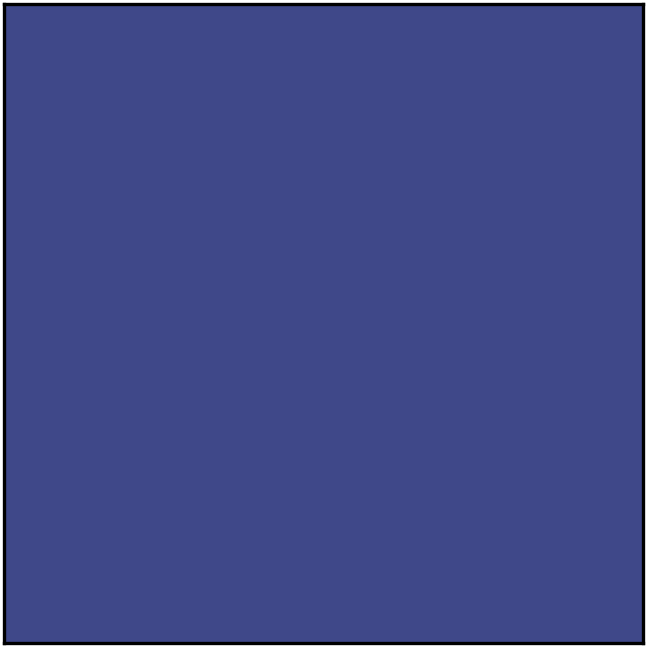} &
        	\includegraphics[width=\plotsizeplane]{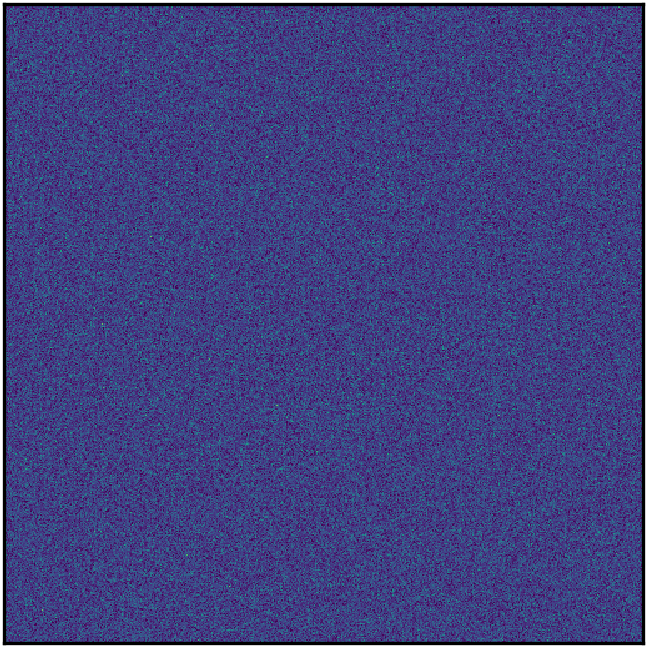} &
        	\includegraphics[width=\plotsizeplane]{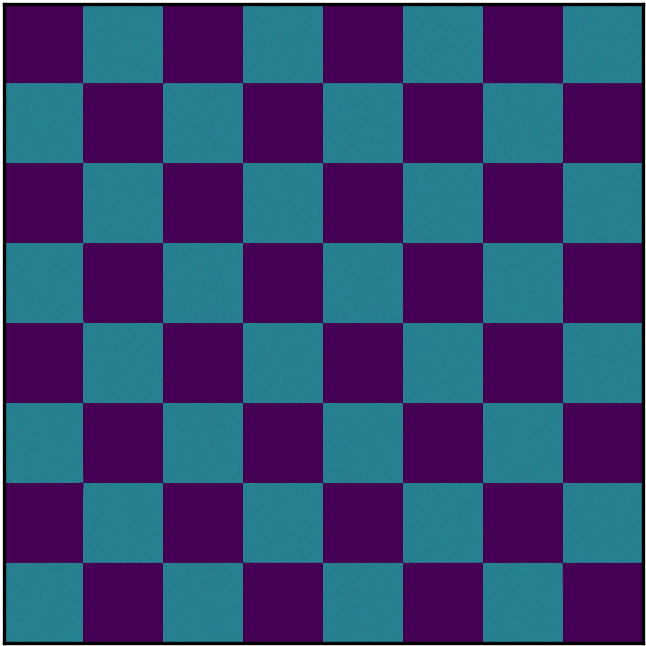} &
        	\includegraphics[width=\plotsizeplane]{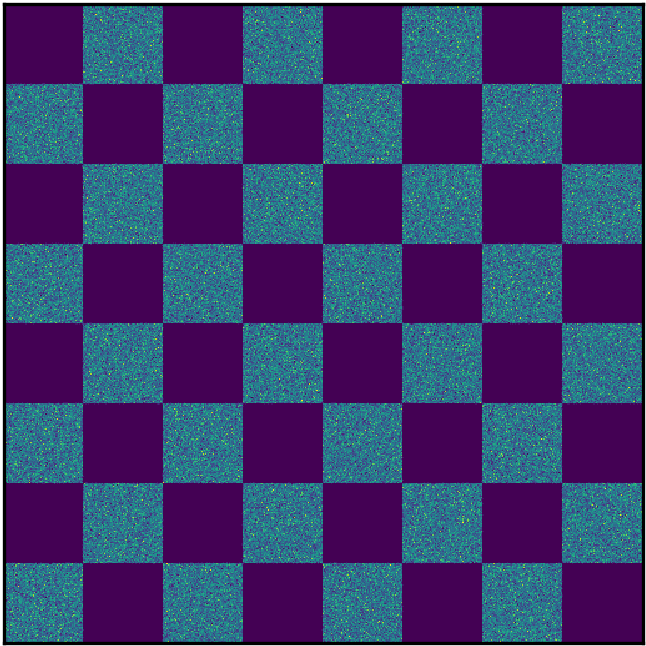}
        	\\
        	\includegraphics[width=\plotsizeplane]{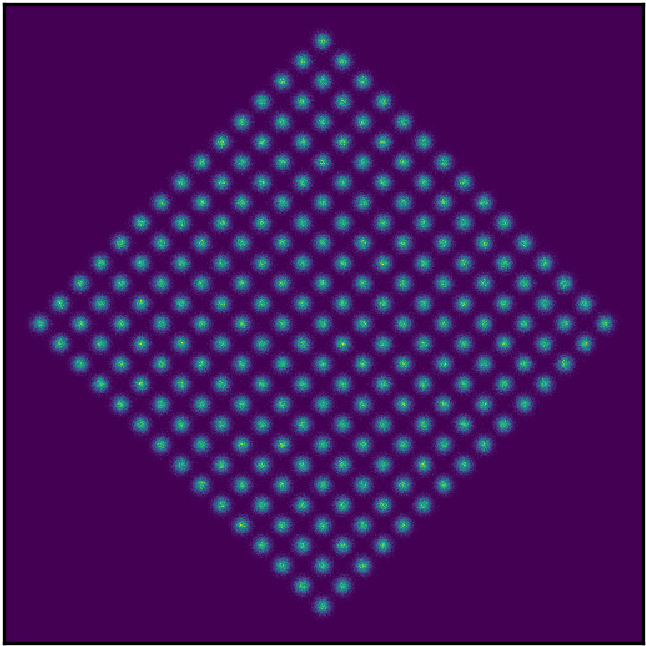} &
        	\includegraphics[width=\plotsizeplane]{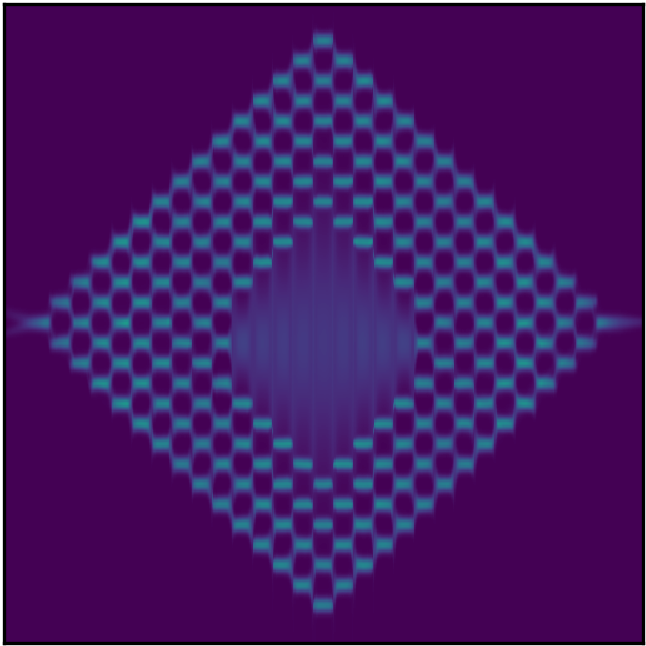} &
        	\includegraphics[width=\plotsizeplane]{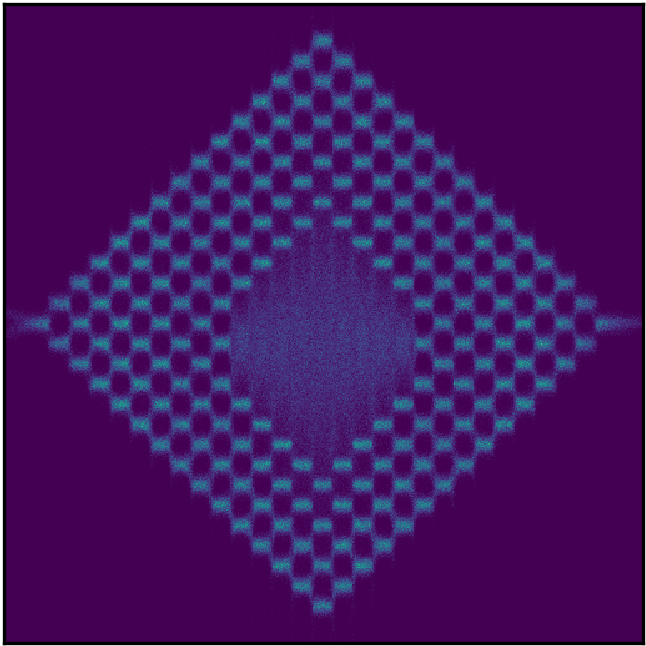} &
        	\includegraphics[width=\plotsizeplane]{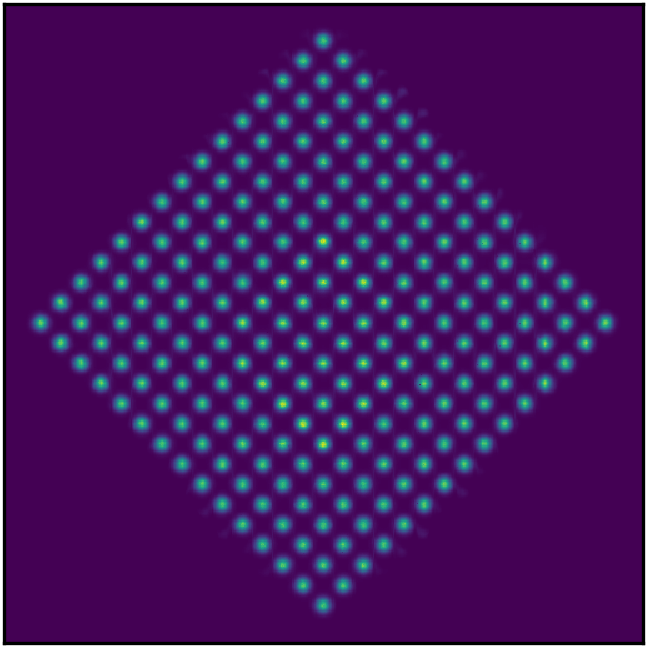} &
        	\includegraphics[width=\plotsizeplane]{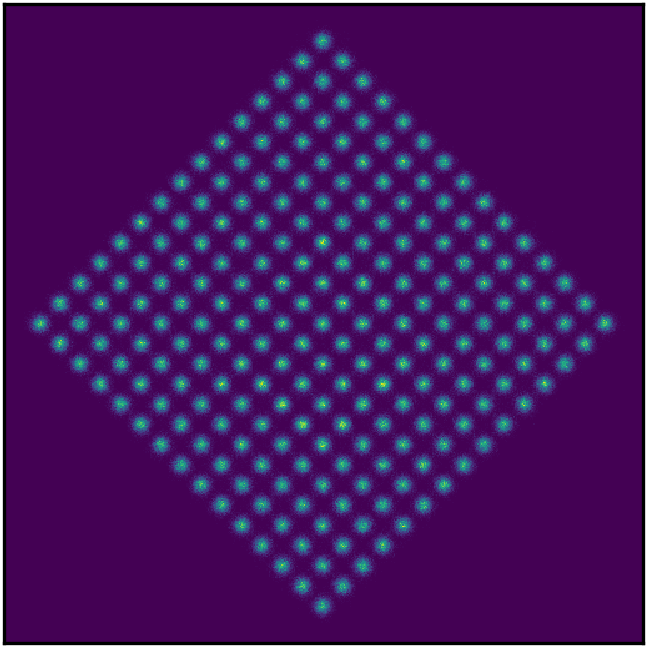}
        	\\
        	\includegraphics[width=\plotsizeplane]{./figures/einstein/einstein-data.png} &
        	\includegraphics[width=\plotsizeplane]{./figures/einstein/einstein-proposal.png} &
        	\includegraphics[width=\plotsizeplane]{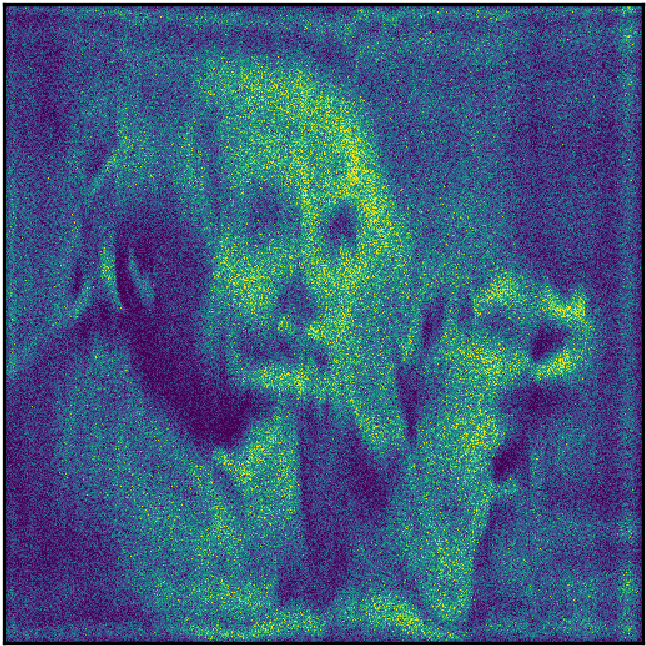} &
        	\includegraphics[width=\plotsizeplane]{./figures/einstein/einstein-aem.png} &
        	\includegraphics[width=\plotsizeplane]{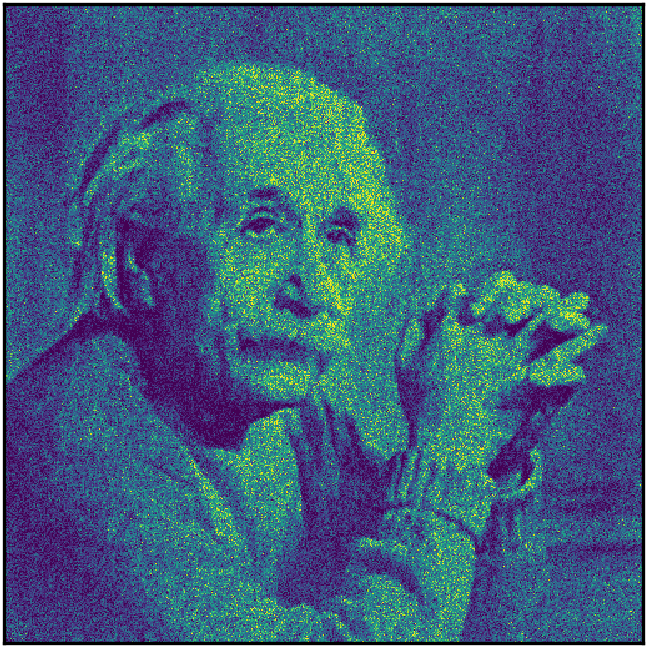}
        \end{tabular}
        \caption{}
        \label{fig:plane-fits}
    \end{subfigure}
    \begin{subfigure}[b]{0.398\textwidth}
        \centering
        \setlength\tabcolsep{0pt}
        \newcolumntype{P}[1]{>{\centering\arraybackslash}p{#1}}
        \begin{tabular}{P{0.8333in}P{0.8333in}P{0.8333in}}
        \multicolumn{3}{c}{Einstein conditionals $ \propto p(x_{2} \vert x_{1}) $} \\
        \cmidrule(lr){1-3}
        Image & Proposal & AEM \\
        \multicolumn{3}{c}{
    	\includegraphics[width=2.5in, height=3.085in]{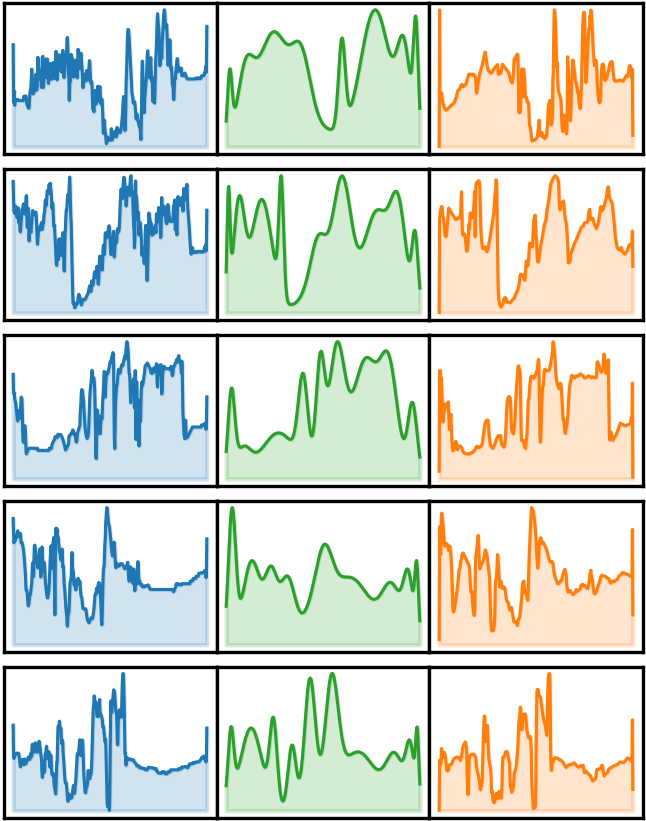}
    	} \\
        \end{tabular}
        \caption{}
        \label{fig:einstein-slices}
    \end{subfigure}
	\caption{(a) Estimated densities and samples for both the proposal distribution and AEM on a range of synthetic two-dimensional densities. For the checkerboard grid (row 2), we used fixed uniform conditionals for the proposal distribution. AEM densities are evaluated with a normalizing constant estimated using 1000 importance samples. We present histograms of $ 10^{6} $ samples from each trained model, for comparison with each training dataset of the same cardinality. (b) Unnormalized conditional distributions proportional to $ p(x_{2} \vert x_{1}) $ for a selection of vertical slices on the Einstein task. An energy-based approach allows for a better fit to the true pixel intensities in the image, which feature high-frequency components ill-suited for a finite mixture of Gaussians.}
\end{figure*}
\renewcommand\arraystretch{1.}

\subsection{Sampling}\label{subsec:sampling}
Although it is not possible to sample analytically from our model, we can obtain approximate samples by first drawing samples from the proposal distribution and then resampling from this collection using importance weights computed by the energy model (eq.~\eqref{eq:importance_weights}). This method is known as sampling importance resampling \cite{rubin1988using}, and is consistent in the limit of infinite proposal samples, but we found results to be satisfactory using just 100 proposal samples. 

\subsection{ResMADE}
For general purpose density estimation of tabular data, we present a modified version of the MADE architecture \cite{germain2015made} that incorporates residual connections \cite{he2016deep}. In the standard MADE architecture, causal structure is maintained by masking certain weights in the network layers. We observe that in consecutive hidden layers with the same number of units and shared masks, the connectivity of the hidden units with respect to the inputs is preserved. As such, incorporating skip connections will maintain the architecture's autoregressive structure. 

Following the documented success of residual blocks \cite{he2016deep, he2016preact} as a component in deep network architectures, we implement a basic residual block for MADE-style models that can be used as a drop-in replacement for typical masked layers.
Each block consists of two masked-dense layers per residual block, and uses pre-activations following \citet{he2016preact}. We use the term ResMADE to describe an autoregressive architecture featuring these blocks, and make use of the ResMADE as an ARNN component across our experiments. For a more detailed description see Appendix \ref{app:resmade}.


\section{Experiments}\label{sec:experiments}
For our experiments, we use a ResMADE with four residual blocks for the ARNN, as well as a fully-connected residual architecture for the ENN, also with four residual blocks. The number of hidden units in the ResMADE is varied per task. We use the Adam optimizer \cite{kingma2014adam}, and anneal the learning rate to zero over the course of training using a cosine schedule \cite{loshchilov2016sgdr}. For some tasks, we find regularization by dropout \cite{srivastava2014dropout} to be beneficial. Full experimental details are available in Appendix \ref{app:experimental}, and code is available at \url{https://github.com/conormdurkan/autoregressive-energy-machines}.

\subsection{Synthetic datasets}
We first demonstrate that an AEM is capable of fitting complex two-dimensional densities. For each task, we generate $ 10^{6} $ data points for training. Results are displayed in \cref{fig:plane-fits}. We plot each AEM density by estimating the normalizing constant with 1000 importance samples for each conditional. AEM samples are obtained by resampling 100 proposal samples as described in \cref{subsec:sampling}.

\textbf{Spirals}
The spirals dataset is adapted from \citet{grathwohl2018ffjord}. Though capable of representing the spiral density, the ResMADE proposal fails to achieve the same quality of fit as an AEM, with notable regions of non-uniform density. 

\textbf{Checkerboard} The checkerboard dataset is also adapted from \citet{grathwohl2018ffjord}. This task illustrates that in some cases a learned proposal distribution is not required; with a fixed, uniform proposal, an AEM is capable of accurately modeling the data, including the discontinuities at the boundary of each square. 

\textbf{Diamond} The diamond task is adapted from the 100-mode square Gaussian grid of \citet{huang2018naf}, by expanding to 225 modes, and rotating 45 degrees. Although the ResMADE proposal does not have the capacity to represent all modes of the target data, an AEM is able to recover these lost modes. 

\textbf{Einstein} We generate the Einstein data by sampling co-ordinates proportional to the pixel intensities in an image of Albert Einstein \cite{muller2018neuralimportancesampling}, before adding uniform noise, and re-scaling the resulting points to $ [0, 1]^{2} $. This task in particular highlights the benefits of an energy-based approach. The distribution of light in an image features sharp transitions and edges, high-frequency components, and broad regions of near-constant density. Where a ResMADE proposal struggles with these challenges, an AEM is able to retain much more fine detail. In addition, samples generated by the AEM are difficult to distinguish from the original dataset. 

Finally, \cref{fig:einstein-slices} presents an alternative visualization of the Einstein task. Each row corresponds to an unnormalized conditional proportional to $ p(x_{2} \vert x_{1}) $ for a fixed value of $ x_{1} $. While the ResMADE proposal, consisting of a mixture of 10 Gaussians for each conditional, achieves a good overall fit to the true pixel intensities, it is ultimately constrained by the smoothness of its mixture components.

\subsection{Normalizing constant estimation}\label{subsec:well-calibrated}
Though importance sampling in one dimension is much less unwieldy than in high-dimensional space, it may still be the case that the proposal distributions do not adequately cover the support of the conditionals being modeled, leading to underestimates of the normalizing constants. Here we demonstrate that the normalizing constants learned by an AEM are well-calibrated by comparing to `true' values computed with explicit numerical integration. In particular, we use a log-modified trapezoidal rule \cite{pitkin2017lintegrate} to integrate the unnormalized log density output by the energy network over each dimension. This approach exploits the fast parallel computation of the energy net, allowing us to saturate the domain of integration, and compensate for the shortcomings of the trapezoidal rule compared to more advanced adaptive quadrature methods \cite{gander2000adaptive}. We increase the number of integrand evaluations until the integral converges to seven significant figures.

\begin{figure}[ht]
	\centering
	\includegraphics[width=3in, height=3.2in]{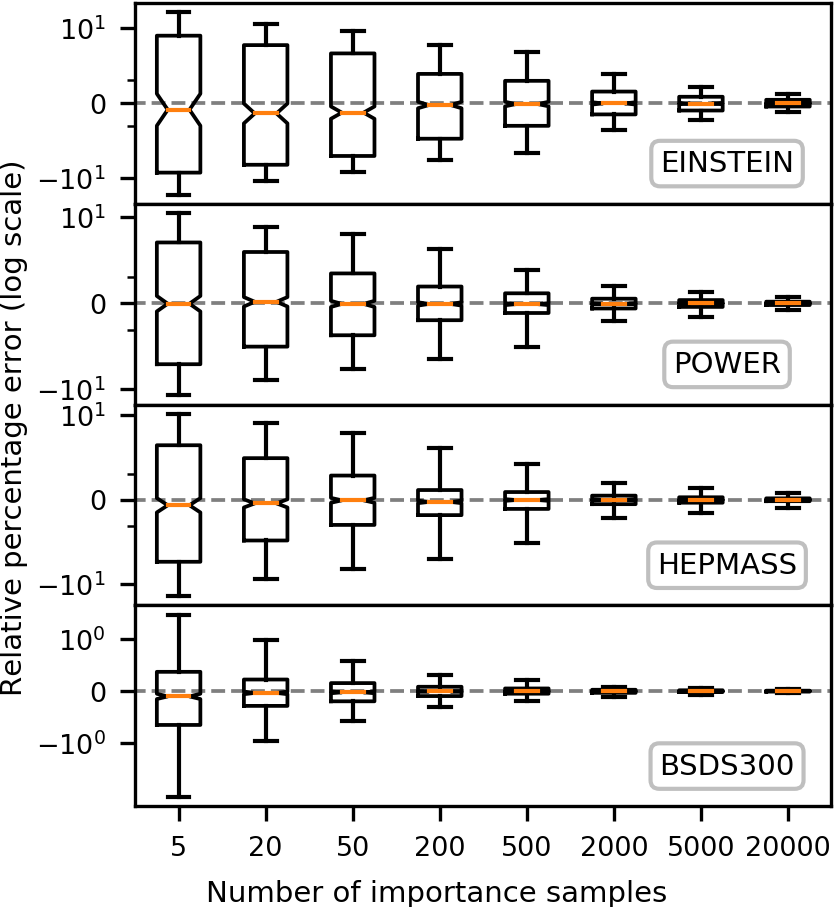}
	\caption{Accuracy of log normalizing constant estimates increases with number of importance samples for each task. Whiskers delineate the $ 5^{th} $ and $ 95^{th} $ percentiles of the relative error. The Einstein conditionals, as illustrated in \cref{fig:einstein-slices}, prove particularly difficult, while BSDS300 conditionals are in contrast much simpler to approximate.}
	\label{normalizing-constants}
\end{figure}

For each trained model on the Einstein, Power, Hepmass, and BSDS300 tasks, we first randomly select 1000 dimensions with replacement according to the data dimensionality, disregarding the marginal $ p(x_{1}) $ so as to not repeatedly compute the same value. Then, we compute the integral of the log unnormalized density corresponding to that one-dimensional conditional, using a log-trapezoidal rule and context vectors generated from a held out-validation set of 1000 samples. This procedure results in 1000 `true' integrals for each task. We then test the AEM by comparing this true value with estimates generated using increasing numbers of importance samples. Note that in each task, the AEM has never estimated the log normalizing constant for the conditional densities under consideration, since these conditionals are specified using the validation set, and not the training set. \Cref{normalizing-constants} visualizes the results of our calibration experiments. 

\begin{table*}[ht]
	\caption{Test log likelihood (in nats) for UCI datasets and BSDS300, with error bars corresponding to two standard deviations. AEM$^*$ results are estimated with 20,000 importance samples. The best performing model for each dataset is shown in bold, as well the best performing model for which the exact log likelihood can be obtained. Results for non-AEM models taken from existing literature. MAF-DDSF$^\dagger$ report error bars across five repeated runs rather than across the test set.}
	\label{tab:uci-results}
	\begin{center}
		\begin{small}
			\begin{sc}
				\begin{tabular}{lccccc}
					\toprule
                    Model                       & POWER           & GAS              & HEPMASS           & MINIBOONE         & BSDS300           \\ \midrule
                    MADE-MoG                    & $0.40 \pm 0.01$ & $8.47 \pm 0.02$ & $-15.15 \pm 0.02$ & $-12.27 \pm 0.47$ & $153.71 \pm 0.28$ \\
                    MAF & $0.30 \pm 0.01$ & $10.08 \pm 0.02$ & $-17.39 \pm 0.02$ & $-11.68 \pm 0.44$ & $156.36 \pm 0.28$ \\
                    MAF-DDSF$^\dagger$ & $0.62 \pm 0.01$ & $11.96 \pm 0.33$ & $-15.09 \pm 0.40$ & $\bm{-8.86 \pm 0.15}$ & $157.73 \pm 0.04$ \\
                    TAN (various) & $0.60 \pm 0.01$ & $12.06 \pm 0.02$ & $-13.78 \pm 0.02$ & $-11.01 \pm 0.48$ & $\bm{159.80 \pm 0.07}$ \\ \midrule
                    ResMADE-MoG (proposal)              &     $0.61 \pm 0.01$            & $12.80 \pm 0.01$                  &     $-13.42 \pm 0.01$               &    $-11.01 \pm 0.23$                 & $157.41 \pm 0.14$                    \\
                    AEM-KDE                         &  $\bm{0.65 \pm 0.01 }$                &     $\bm{12.89 \pm 0.01}$              &    $\bm{-12.87 \pm 0.01}$                &     $-10.33 \pm 0.22$                &    $158.44 \pm 0.14$               \\ 
                    AEM$^*$             & $\bm{0.70 \pm 0.01}$                  &      $\bm{13.03 \pm 0.01}$            &     $\bm{-12.85 \pm 0.01}$               &     $-10.17 \pm 0.26$                &  $158.7 1 \pm 0.14$ \\ \bottomrule
				\end{tabular}
			\end{sc}
		\end{small}
	\end{center}
\end{table*}

\subsection{Density estimation on tabular data}
We follow the experimental setup of \citet{papamakarios2017maf} in using a selection of pre-processed datasets from the UCI machine learning repository \cite{dua2017uci}, and BSDS300 datasets of natural images \cite{martin2001bsds}. AEM log likelihoods are estimated using 20000 importance samples from the proposal distribution.

As a normalized approximation to the AEM, we use a kernel density estimate in which Gaussian kernels are centered at proposal samples and weighted by the importance weights. We include the proposal itself as a mixture component in order to provide probability density in regions not well-covered by the samples. The KDE bandwidth and proposal distribution mixture weighting are optimized on the validation set. We call the model under this evaluation scheme AEM-KDE\@. KDE estimates also use 20000 samples from the proposal distribution for each conditional.

\Cref{tab:uci-results} shows the test-set log likelihoods obtained by our models and by other state-of-the-art models. We first note that the AEM proposal distribution (ResMADE-MoG) provides a strong baseline relative to previous work. In particular, it improves substantially on the the MADE-MoG results reported by \citet{papamakarios2017maf}, and improves on the state-of-the-art results reported by NAF \cite{huang2018naf} and TAN \cite{oliva2018tan}. This demonstrates the benefits of adding residual connections to the MADE architecture. As such, practitioners may find ResMADE a useful component in many applications which require autoregressive computation. 

\begin{table}[t]
    \caption{Latent variable modeling results with AEM priors. AEM-VAE$^{\star}$ results obtained with 1000 importance samples and $\log p(\bfx)$ lower-bounded using the method of \citet{burda2015importance} with 50 samples. IAF-DSF$^\dagger$ report error bars across five repeated runs rather than across the test set.}
    \label{tab:vae}
    \centering
    \begin{tabular}{lcc}
    \toprule
    Model       & ELBO              & $\log p(\bfx)$    \\ \midrule
    
    IAF-DSF$^\dagger$     & $-81.92 \pm 0.04$ & $-79.86 \pm 0.01$ \\
    VAE         & $-84.43 \pm 0.23$ & $-81.23 \pm 0.21$ \\
    ResMADE-VAE & $-82.96 \pm 0.23$ & $-79.89 \pm 0.21$ \\
    AEM-KDE-VAE & $-82.95 \pm 0.23$ & $-79.88 \pm 0.21$ \\
    AEM-VAE$^{\star}$     & $-82.92 \pm 0.23$ & $-79.87 \pm 0.21$ \\ \bottomrule
    \end{tabular}
\end{table}

AEM and AEM-KDE outperform both the proposal distribution and existing state-of-the-art methods on the Power, Gas and Hepmass datasets, demonstrating the potential benefit of flexible energy-based conditionals. Despite regularization, overfitting was an issue for Miniboone due to the size of the training set ($ n=29,556 $) relative to the data dimension ($ D = 43 $). This highlights the challenges associated with using very expressive models on domains with limited data, and the need for stronger regularization methods in these cases. On BSDS300, our models achieve the second highest scores relative to previous work. On this dataset we found that the validation scores ($\sim 174$) were substantially higher than test-set scores ($\sim 158$), indicating differences between their empirical distributions. Overall, the AEM-KDE obtains improved scores relative to the proposal distribution, and these scores are close to those of the AEM.

\subsection{Latent variable modeling}
We evaluate the AEM in the context of deep latent-variable models, where it can be used as an expressive prior. We train a convolutional variational autoencoder (VAE) \cite{kingma2013vae, rezende2014stochastic}, making use of residual blocks in the encoder and decoder in a similar architecture to previous work \cite{huang2018naf, kingma2016iaf}. We initially train the encoder and decoder with a standard Gaussian prior, and then train an AEM post-hoc to maximize the likelihood of samples from the aggregate approximate posterior 
\begin{align}
    q_{\text{agg}}(\bfz) = \mathbb{E}_{\bfx \sim p^{\star}(\bfx)}[q(\bfz | \bfx)].
\end{align}
This training method avoids an issue where maximum-likelihood training of the proposal distribution interferes with the lower bound objective. During pre-training, we set $\beta = 0.9$ in the modified variational objective
\begin{align}
    \mathcal{L}_\beta = \mathbb{E}_{q(\bfz | \bfx)}[\log p(\bfx | \bfz)] - \beta \infdiv{q(\bfz | \bfx)}{p(\bfz)}.
\end{align}
This weighting of the KL-divergence term has the effect of boosting the reconstruction log probability at the expense of an aggregate posterior that is less well-matched to the standard Gaussian prior. This provides an opportunity for the AEM to improve on the original prior as a model of the aggregate posterior. Weighting of the KL-divergence has been used in previous work to reduce the occurrence of unused latent variables \cite{sonderby2016ladder}, and to control the types of representations encoded by the latent variables \cite{higgins2016beta}.

\begin{figure}[ht]
	\centering
	\begin{subfigure}{0.49\columnwidth}
    	\centering
	    \includegraphics[width=\columnwidth]{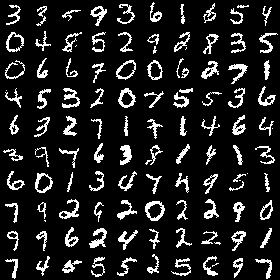}
	    \caption{Data}
	\end{subfigure}
	\begin{subfigure}{0.49\columnwidth}
	    \centering
	    \includegraphics[width=\columnwidth]{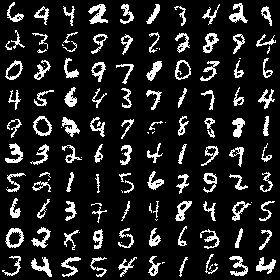}
	    \caption{AEM-VAE}
	\end{subfigure}
	\caption{Binarized MNIST data examples and unconditional samples for AEM-VAE obtained by resampling from 100 proposal samples.}
	\label{fig:mnist}
\end{figure}

\Cref{tab:vae} shows that the AEM-VAE improves substantially on the standard Gaussian prior, and that the results are competitive with existing approaches on dynamically-binarized MNIST \cite{burda2015importance}. The AEM does not improve on the proposal distribution scores, possibly because the aggregate latent posterior is a mixture of Gaussians, which is well-modeled by the ResMADE-MoG\@. Figure \ref{fig:mnist} shows data examples and samples from the trained model.

\section{Related Work}\label{sec:related-work}

\textbf{Flow-based neural density estimation}
Together with autoregressive approaches, flow-based models have also seen widespread use in neural density estimation. Flow-based models consist of invertible transformations for which Jacobian determinants can be efficiently computed, allowing exact density evaluation through the change-of-variables formula. Multiple transformations are often stacked to enable more complex models.

\textbf{Efficiently invertible flows}
Flows exploiting a particular type of transformation, known as a coupling layer, not only allow for one-pass density evaluation, but also one-pass sampling. Examples of such models include NICE \cite{dinh2014nice} and RealNVP \cite{dinh2016realnvp}, and the approach has recently been extended to image \cite{kingma2018glow} and audio \cite{prenger2018waveglow, kim2018flowavenet} data. The case of continuous flows based on ordinary differential equations has also recently been explored by FFJORD \cite{grathwohl2018ffjord}. However, efficient sampling for this class of models comes at the cost of density estimation performance, with autoregressive models generally achieving better log likelihood scores.

\textbf{Autoregressive flows} Originally proposed by \citet{kingma2016iaf} for variational inference, autoregressive flows were adapted for efficient density estimation by \citet{papamakarios2017maf} with MAF. Subsequent models such as NAF \cite{huang2018naf} and TAN \cite{oliva2018tan} have developed on this idea, reporting state-of-the-art results for density estimation. Sampling in these models is expensive, since autoregressive flow inversion is inherently sequential. In some cases, such as \citet{huang2018naf}, the flows do not have an analytic inverse, and must be inverted numerically for sampling. Despite these caveats, autoregressive density estimators remain the best performing neural density estimators for general density estimation tasks.

\textbf{Energy-based models} In this work we describe energy-based models as unnormalized densities that define a probability distribution over random variables. However, there exist multiple notions of energy-based learning in the machine learning literature, including non-probabilistic interpretations \cite{lecun2006tutorial, zhao2016energy}. We focus here on recent work which includes applications to density estimation with neural energy functions. Deep energy estimator networks \cite{saremi2018deen} use an energy function implemented as a neural network, and train using the score-matching framework. This objective avoids the need to estimate the normalizing constant, but also makes it challenging to compare log-likelihood scores with other density estimators. \citet{bauer2018resampled} propose an energy-based approach for increasing the flexibility of VAE priors, in which a neural network energy function is used to mask a pre-specified proposal function. As in our work, training is performed using importance sampling, but due to the larger dimensionality of the problem, $2^{10}$ samples were used in the estimates during training.

\textbf{Other related work} \citet{muller2018neuralimportancesampling} propose neural importance sampling, in which a flow-based neural sampler is optimized in order to perform low-variance Monte Carlo integration of a given target function. This is similar to the goal of the proposal distribution in the AEM, but in our case the proposal is trained jointly with an energy model, and we do not assume that the target function is known a priori.


\section{Conclusion}\label{sec:conclusion}

We proposed the Autoregressive Energy Machine, a neural density estimator that addresses the challenges of energy-based modeling in high dimensions through a scalable and efficient autoregressive estimate of the normalizing constant. While exact density evaluation is intractable for an AEM, we have demonstrated that the flexibility of an energy-based model enables us to model challenging synthetic data, as well as achieve state-of-the-art results on a suite of benchmark datasets.




\newpage
\section*{Acknowledgements}
The authors thank George Papamakarios, Iain Murray, and Chris Williams for helpful discussion. This work was supported in part by the EPSRC Centre for Doctoral Training in Data Science, funded by the UK Engineering and Physical Sciences Research Council (grant EP/L016427/1) and the University of Edinburgh. 
\bibliographystyle{icml2019}
\bibliography{main.bbl}

\newpage
\appendix
\section{ResMADE}\label{app:resmade}
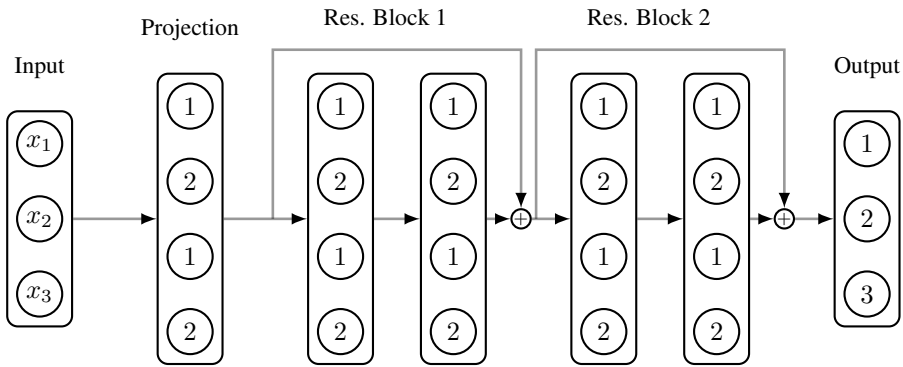
\begin{figure*}[t]
	\centering
{
    \begin{tikzpicture}[auto, thick, node distance=1cm]
    
        \draw
        node [var] (x_1) {$x_1$}
    	node [var, below of=x_1] (x_2) {$x_2$}
    	node [var, below of=x_2] (x_3) {$x_3$}
    	node [block, fit=(x_1)(x_3), label={[yshift=0.3cm]Input}] (input-block) {};
    	
        \draw
        node [var, right of=x_1, yshift=0.5cm, xshift=1cm] (p_1) {$1$}
    	node [var, below of=p_1] (p_2) {$2$}
    	node [var, below of=p_2] (p_3) {$1$}
    	node [var, below of=p_3] (p_4) {$2$}
    	node [block, fit=(p_1)(p_4), label={[yshift=0.3cm]Projection}] (proj-block) {};
    	
        \coordinate[right of=p_2, xshift=0.1cm, yshift=-0.5cm] (invis_1);
    	
    
        \draw
        node [var, right of=p_1, xshift=1cm] (r_11) {$1$}
    	node [var, below of=r_11] (r_12) {$2$}
    	node [var, below of=r_12] (r_13) {$1$}
    	node [var, below of=r_13] (r_14) {$2$}
    	node [block, fit=(r_11)(r_14), label={[yshift=0.5cm, xshift=0.6cm]Res. Block 1}] (res-block-11) {}
    	
        node [var, right of=r_11, xshift=0.5cm] (r_21) {$1$}
    	node [var, below of=r_21] (r_22) {$2$}
    	node [var, below of=r_22] (r_23) {$1$}
    	node [var, below of=r_23] (r_24) {$2$}
    	node [block, fit=(r_21)(r_24)] (res-block-12) {};
    	
    	\draw
        node [var, right of=r_22, xshift=-0.1cm, yshift=-0.5cm, inner sep=0pt, minimum size=1pt] (plus_1) {\tiny $+$};
    	
        \draw
        node [var, right of=r_21, xshift=1cm] (r_31) {$1$}
    	node [var, below of=r_31] (r_32) {$2$}
    	node [var, below of=r_32] (r_33) {$1$}
    	node [var, below of=r_33] (r_34) {$2$}
    	node [block, fit=(r_31)(r_34), label={[yshift=0.5cm, xshift=0.6cm]Res. Block 2}] (res-block-21) {}
    	
        node [var, right of=r_31, xshift=0.5cm] (r_41) {$1$}
    	node [var, below of=r_41] (r_42) {$2$}
    	node [var, below of=r_42] (r_43) {$1$}
    	node [var, below of=r_43] (r_44) {$2$}
    	node [block, fit=(r_41)(r_44)] (res-block-22) {};
    	
    	\draw
        node [var, right of=r_42, xshift=-0.1cm, yshift=-0.5cm, inner sep=0pt, minimum size=1pt] (plus_2) {\tiny $+$};
        
        \coordinate[right of=r_22, xshift=0.1cm, yshift=-0.5cm] (invis_2);
    	
        \draw
        node [var, right of=r_41, yshift=-0.5cm, xshift=1cm] (o_1) {$1$}
    	node [var, below of=o_1] (o_2) {$2$}
    	node [var, below of=o_2] (o_3) {$3$}
    	node [block, fit=(o_1)(o_3), label={[yshift=0.3cm]Output}] (out-block) {};
    	
    	\draw[customline, ->] (input-block.east) -- (proj-block.west);
    	\draw[customline, ->] (proj-block.east) -- (res-block-11.west);
    	\draw[customline, ->] (res-block-11.east) -- (res-block-12.west);
    	\draw[customline, ->] (res-block-12.east) -- (plus_1);
    	\draw[customline, ->] (plus_1) -- (res-block-21.west);
    	\draw[customline, ->] (res-block-21.east) -- (res-block-22.west);
    	\draw[customline, ->] (res-block-22.east) -- (plus_2);
    	\draw[customline, ->] (plus_2) -- (out-block.west);
    	
    	\draw[customline, ->] (invis_1) --  ++(0,2.25) -| (plus_1);
    	\draw[customline, ->] (invis_2) --  ++(0,2.25) -| (plus_2);

    
    
    	
    	
    \end{tikzpicture}
}
	\caption{ResMADE architecture with $ D = 3 $ input data dimensions and $ H = 4 $ hidden units. The degree of each hidden unit and output is indicated with an integer label. Sequential degree assignment results in each hidden layer having the same masking structure, here alternating between dependence on the first input, or the first two inputs. These layers can be combined using any binary elementwise operation, while preserving autoregressive structure. In particular, residual connections can be added in a straightforward manner.The ResMADE architecture consists of an initial masked projection to the target hidden dimensionality, a sequence of masked residual blocks, and finally a masked linear layer to the output units.}
	\label{fig:resmade}
\end{figure*}

Residual connections \cite{he2016deep} are widely used in deep neural networks, and have demonstrated favourable performance relative to standard networks. Residual networks typically consist of many stacked transformations of the form
\begin{align}
    \bfh_{l + 1} = \bfh_l + \bff({\bfh_l}),
\end{align}
where $\bff$ is a residual block. In this work, we observe that it is possible to equip a MADE \cite{germain2015made} with residual connections when certain conditions on its masking structure are met. 

At initialization, each unit in each hidden layer of a MADE is assigned a positive integer, termed its degree. The degree specifies the number of input dimensions to which that particular unit is connected. For example, writing $ d_{k}^{l} $ for unit $ k $ of layer $ l $, a value $ d_{k}^{l} = m $ means that the unit depends only on the first $ m $ dimensions of the input, when the input is prescribed a particular ordering (we always assume the ordering given by the data). In successive layers, this means that a unit may only be connected to units in the previous layer whose degrees strictly do not exceed its own. In other words, the degree assignment defines a binary mask matrix which multiplies the weight matrix of each layer in a MADE elementwise, maintaining autoregressive structure. Indeed, the mask $ M $ for layer $ l $ is given in terms of the degrees for layer $ l - 1 $ and layer $ l $: 
\begin{align}
    M_{ij}^{l} = 
    \begin{cases}
        1 \hspace{0.25cm} \text{if $ d_{i}^{l} \geq d_{j}^{l - 1} $ }\\
        0 \hspace{0.25cm} \text{otherwise}.
    \end{cases}
\end{align}

Through this masking process, it can be guaranteed that output units associated with data dimension $d$ only depend on the previous inputs $\bfx_{<d}$. Though the original MADE paper considered a number of ways in which to assign degrees, we focus here on fixed sequential degree assignment, used also by MAF \cite{papamakarios2017maf}. 
For an input of dimension $ D $, we define the degree
\begin{align}
    d_{k}^{l} = (k - 1) \ (\mathrm{mod}\ (D - 1)) + 1.
\end{align}
We also assume the dimension $ H $ of each hidden layer is at least $ D $, so that no input information is lost. The result of sequential degree assignment for $ D = 3 $ and $ H = 4 $ is illustrated in \cref{fig:resmade}.

If all hidden layers in the MADE are of the same dimensionality $ H $, sequential degree assignment means that each layer will also have the same degree structure. In this way, two vectors of hidden units in the MADE may be combined using any binary element-wise operation, also shown in Figure \cref{fig:resmade}. In particular, a vector computed from a fully-connected block with masked layers can be added to the input to that block while maintaining the same autoregressive structure, allowing for the traditional residual connection to be added to the MADE architecture. 

Not only do residual connections enhance a MADE in its own right, but the resulting ResMADE architecture can be used as a drop-in replacement wherever a MADE is used as a building block, such as IAF, MAF, or NAF \cite{kingma2016iaf, papamakarios2017maf, huang2018naf}. 
\section{Experimental settings}\label{app:experimental}
In all experiments, we use the same ResMADE and ENN architectures, each with 4 pre-activation residual blocks \cite{he2016preact}. The number of hidden units and the dimensionality of the context vector for the ENN are fixed across all tasks at 128 and 64 respectively. The number of hidden units in the ResMADE is tuned per experiment. For the exact experimental settings used in each experiment, see Tables \ref{tab:plane-vae-settings} and \ref{tab:uci-settings}.

For the proposal distributions, we use a mixture of Gaussians in all cases except for the checkerboard experiment, where we use a fixed uniform distribution. We use 10 mixture components for the synthetic experiments, and 20 components for all other experiments. We use a minimum scale of $ 10^{-3}$ for the Gaussian distributions in order to prevent numerical issues.

For optimization we use Adam \cite{kingma2014adam} with a cosine annealing schedule \cite{loshchilov2016sgdr} and an initial learning rate of $5 \times 10^{-4}$. The number of training steps is adjusted per task. For Miniboone, we use only 6000 training updates, as we found overfitting to be a significant problem for this dataset. Early-stopping is used to select models, although in most cases the best models are obtained at the end of training. Normalizing constants are estimated using 20 importance samples, and dropout is applied to counter overfitting, with a rate that is tuned per task. Dropout is applied between the two layers of each residual block in both the ENN and ResMADE. 

For the VAE, we use an architecture similar to those used in IAF and NAF \cite{kingma2016iaf, huang2018naf}. For the encoder, we alternate between residual blocks that preserve spatial resolution, and downsampling residual blocks with strided convolutions. The input is projected to 16 channels using a $1 \times 1$ convolution, and the number of channels is doubled at each downsampling layer. After three downsampling layers we flatten the feature maps and use a linear layer to regress the means and log-variances of the approximate posterior with 32 latent units. The decoder mirrors the encoder, with the input latents being linearly projected and reshaped to a $ [4, 4, 128] $ spatial block, which is then upsampled using transpose-convolutions in order to output pixel-wise Bernoulli logits. 

\begin{table*}[h!]
	\caption{Experimental setting for synthetic data and VAE experiments.}
	\label{tab:plane-vae-settings}
	\begin{center}
		\begin{small}
			\begin{sc}
				\begin{tabular}{lccccc}
					\toprule
                    Hyperparameter                       & SPIRALS           & CHECKERBOARD              & DIAMOND           &  EINSTEIN         & VAE          \\ \midrule
                    Batch size & 256 & 256 & 256 & 256 & 256 \\
                    ResMADE hidden dim. & 256 & 256 & 256 & 256 & 512 \\
                    ResMADE activation & ReLU & ReLU & ReLU & ReLU & ReLU \\
                    ResMADE dropout & 0 & 0 & 0 & 0 & 0.5 \\
                    Context dim. & 64 & 64 & 64 & 64 & 64 \\
                    ENN hidden dim. & 128 & 128 & 128 & 128 & 128 \\
                    ENN activation & ReLU & ReLU & ReLU & ReLU & ReLU \\
                    ENN dropout & 0 & 0 & 0 & 0 & 0.5 \\
                    Mixture comps. & 10 & - & 10 & 10 & 20 \\
                    Mixture comp. scale min. & 1E-3 & - & 1E-3 & 1E-3 & 1E-3 \\ 
                    Learning rate & 5e-4 & 5e-4 & 5e-4 & 5e-4 & 5e-4 \\
                    Total steps & 400000 & 400000 & 400000 & 3000000 & 100000 \\
                    Warm-up steps & 5000 & 0 & 0 & 0 & 0 \\
                     \bottomrule
				\end{tabular}
			\end{sc}
		\end{small}
	\end{center}
\end{table*}

\begin{table*}[h]
	\caption{Experimental settings for UCI and BSDS300 datasets.}
	\label{tab:uci-settings}
	\begin{center}
		\begin{small}
			\begin{sc}
				\begin{tabular}{lccccc}
					\toprule
                    Hyperparameter                       & POWER           & GAS              & HEPMASS           & MINIBOONE         & BSDS300          \\ \midrule
                    Batch size & 512 & 512 & 512 & 512 & 512 \\
                    ResMADE hidden dim. & 512 & 512 & 512 & 512 & 1024 \\
                    ResMADE activation & ReLU & ReLU & ReLU & ReLU & ReLU \\
                    ResMADE dropout & 0.1 & 0 & 0.2 & 0.5 & 0.2 \\
                    Context dim. & 64 & 64 & 64 & 64 & 64 \\
                    ENN hidden dim. & 128 & 128 & 128 & 128 & 128 \\
                    ENN activation & ReLU & Tanh & ReLU & ReLU & ReLU \\
                    ENN dropout & 0.1 & 0 & 0.2 & 0.5 & 0.2 \\
                    Mixture comps. & 20 & 20 & 20 & 20 & 20 \\
                    Mixture comp. scale min. & 1E-3 & 1E-3 & 1E-3 & 1E-3 & 1E-3 \\ 
                    Learning rate & 5e-4 & 5e-4 & 5e-4 & 5e-4 & 5e-4 \\
                    Total steps & 800000 & 400000 & 400000 & 6000 & 400000 \\
                    Warm-up steps & 5000 & 5000 & 5000 & 0 & 5000 \\
                     \bottomrule
				\end{tabular}
			\end{sc}
		\end{small}
	\end{center}
\end{table*}

\end{document}